\documentclass[letterpaper]{article} 
\usepackage{aaai2026}  
\usepackage{times}  
\usepackage{helvet}  
\usepackage{courier}  
\usepackage[hyphens]{url}  
\usepackage{graphicx} 
\usepackage{natbib}  
\usepackage{caption} 
\usepackage{algorithm}
\usepackage{algorithmic}

\usepackage{newfloat}
\usepackage{listings}

\usepackage{amsmath}
\usepackage{amssymb}
\usepackage{amsfonts}
\usepackage{url}
\usepackage{colortbl}
\usepackage{booktabs}
\usepackage{multirow}
\usepackage{subcaption}
\usepackage[dvipsnames]{xcolor}
\usepackage{transparent}
\usepackage{outlines}
\usepackage{adjustbox}

\usepackage{tcolorbox}

\newcommand{\hjk}[1]{{\color[rgb]{0.0,0.0,0.0}#1}}

\DeclareCaptionStyle{ruled}{labelfont=normalfont,labelsep=colon,strut=off} 
\floatstyle{ruled}
\newfloat{listing}{tb}{lst}{}
\floatname{listing}{Listing}
\title{Improving Large Molecular Language Model\\via Relation-aware Multimodal Collaboration}
\author {
    Jinyoung Park\footnote{Work was done at Korea University.},
    Minseong Bae$^*$,
    Jeehye Na$^*$,
    Hyunwoo J. Kim\footnote{Corresponding Author.}
}
\affiliations {
    Korea Advanced Institute of Science and Technology\\
    \{jinyoung.park, bms2002, jeehyena, hyunwoojkim\}@kaist.ac.kr
}

\usepackage{bibentry}

\begin{document}

\maketitle

\begin{abstract}
Large language models~(LLMs) have demonstrated their instruction-following capabilities and achieved powerful performance on various tasks. 
Inspired by their success, recent works in the molecular domain have led to the development of large molecular language models~(LMLMs) that integrate 1D molecular strings or 2D molecular graphs into the language models.
However, existing LMLMs often suffer from hallucination and limited robustness, largely due to inadequate integration of diverse molecular modalities such as 1D sequences, 2D molecular graphs, and 3D conformations.
To address these limitations, we propose CoLLaMo, a large language model-based molecular assistant equipped with a multi-level molecular modality-collaborative projector.
The relation-aware modality-collaborative attention mechanism in the projector facilitates fine-grained and relation-guided information exchange between atoms by incorporating 2D structural and 3D spatial relations.
Furthermore, we present a molecule-centric new automatic measurement, including a hallucination assessment metric and GPT-based caption quality evaluation to address the limitations of token-based generic evaluation metrics~(\textit{i.e.}, BLEU) widely used in assessing molecular comprehension of LMLMs.  
Our extensive experiments demonstrate that our CoLLaMo enhances the molecular modality generalization capabilities of LMLMs, achieving the best performance on multiple tasks, including molecule captioning, computed property QA, descriptive property QA, motif counting, and IUPAC name prediction.
\end{abstract}
\begin{links}
    \link{Code}{https://github.com/mlvlab/CoLLaMo}
\end{links}

\section{Introduction}
Large language models~(LLMs)~\cite{touvron2023llama,achiam2023gpt,team2023gemini,liu2024visual,dai2023instructblip} have shown their significant advancements in complex reasoning tasks and visual tasks.
Inspired by the progress of the LLMs, growing efforts have been made to adapt language modeling to diverse domains such as table understanding~\cite{zhang2024tablellama,chen2024tablerag,kim2026tabflash}, arithmetic reasoning~\cite{shao2024deepseekmath,yang2024qwen2,chu2025presto}, and scientific domains, particularly molecular modeling~\cite{taylor2022galactica,pei2023biot5}.
Large molecular language models~(LMLMs)~\cite{liu2023molca,li2024towards,yu2024llasmol,park2024llamo} have recently been explored to harness the instruction-following capabilities of LLMs and shown their effectiveness in diverse applications, such as analyzing molecule structures and drug-related tasks.

Despite these promising efforts, current LMLMs face two key challenges. First, they often produce hallucinated outputs, 
especially when tasked with complex molecule-to-language generation. Second, they struggle to consistently leverage the complementary strengths of 1D, 2D, and 3D molecular modalities, which encode distinct but interrelated molecular information.
For instance, 1D SELFIES represent atom sequences, 2D molecular graphs capture topological connectivity, and 3D conformations provide spatial arrangements critical for reasoning about molecular function.
However, existing models typically process these modalities in isolation or shallowly fuse them, limiting their ability to perform molecular reasoning.

To better understand this limitation, we analyze the relationships between molecular tasks and modality contributions.
Our empirical study reveals a modality-task dependency, where the usefulness of each modality varies by task, and no single modality universally dominates.
These findings motivate the need for a unified framework that can collaboratively integrate all molecular modalities while preserving their structural and spatial nuances.

In this work, we propose CoLLaMo, a large language model-based molecular assistant with a modality-collaborative projector that enables joint reasoning over 1D, 2D, and 3D molecular representations.
Specifically, CoLLaMo consists of modality-specific encoders, a large language model, and a modality-collaborative projector that maps representations from different molecular modalities into shared token space through a cross-attention module.
To reflect the fine-grained and coherent information of 2D and 3D modalities in the modality-collaborative projector, we present a relation-aware modality-collaborative attention module, which guides the attention map based on 2D structural and 3D structural proximities.

We also present molecule-centric evaluation measurements to address the limitation of evaluation metrics of LMLMs.
Most prior works rely on generic token-level metrics (e.g., BLEU), which are molecule-agnostic and fail to assess chemical correctness or semantic plausibility.
To this end, we propose a molecule-centric evaluation framework, including (i) a hallucination assessment metric that quantifies factual inconsistencies in model outputs and (ii) a GPT-based caption quality evaluator that assesses molecular descriptions considering molecular information.

Our extensive experiments show that our CoLLaMo consistently outperforms both LMLMs and LLMs, including GPT-based models such as \texttt{GPT-4o}, \texttt{o1-mini} across a wide range of tasks such as molecule captioning, hallucination assessment, IUPAC naming, motif counting, and molecular QA tasks.
Notably, we demonstrate that CoLLaMo maintains strong performance even when certain molecular modalities are missing during inference, showing robustness.
Furthermore, our qualitative analysis highlights that CoLLaMo effectively collaborates representations from diverse molecular modalities and well addresses the hallucination by leveraging 1D, 2D, and 3D information, respectively.

Our contributions are summarized as
\begin{itemize}
    \item We propose a modality-collaborative projector equipped with relation-aware modality-collaborative attention, which facilitates relation-guided information exchange by integrating 2D structural and 3D spatial relations between atoms.
    \item We introduce CoLLaMo, a large molecular language model based on a modality-collaborative projector, which integrates molecular multimodal representations~(1D SELFIES, 2D graphs, and 3D conformations) into unified molecule tokens to fully leverage diverse aspects of molecule information, leading to significantly improved robustness to the molecule hallucination.
    \item We present a molecule-centric evaluation framework for LMLMs, including an automatic hallucination assessment metric and a GPT-based caption quality evaluator, addressing the limitations of conventional token-based metrics (\textit{e.g.}, BLEU).
    \item Our experimental results show that CoLLaMo outperforms the \hjk{strong} baselines including GPT-based models \texttt{GPT-4}, \texttt{GPT-4o}, and \texttt{o1-mini} and large molecular language models across various molecule tasks.
\end{itemize}
\section{Related Works}
\noindent\textbf{Multimodal Large Language Models.}
Multimodal Large Language Models (MLLMs)~\citep{instructblip,liu2024visual,li2023m,yuan2024osprey,liu2024improved} have demonstrated impressive performance on various tasks.
However, since most MLLMs depend on a single vision encoder, they suffer from visual hallucinations~\cite {li2023evaluating,tong2024eyes}.
To address it, multi-encoder strategies have recently been explored, resulting in improved performance and generalization ability ~\citep{lin2023sphinx,kar2024brave, shi2024eagle, fan2024mousi, zong2024mova,shen2024mome,park2025deepvideo}.
Inspired by these findings, we study the impact of individual molecular modalities on large molecular language models and design a molecular modality unified large molecular language model.

\noindent\textbf{Large Molecular Language Models.}
Inspired by the success of CLIP~\citep{radford2021learning}, several studies~\citep{edwards2021text2mol, su2022molecular, liu2023multi, lee2024molecule} have applied molecule-text contrastive learning scheme. 
Furthermore, multiple studies have adopted LLMs~\cite{zeng2022deep,edwards2022translation,pei2023biot5,fang2023mol} to understand molecules using the 1D SMILES or SELFIES.
Recent works have integrated the language models with 2D graph encoders~\cite{kipf2016semi,xu2018powerful} or 3D encoders~\cite{zhou2023uni,zhao2021point,park2023self} with projectors~\cite{zhao2023gimlet,liu2023molca,cao2024presto,park2024llamo}.
MolCA~\citep{liu2023molca}, InstructMol~\citep{cao2023instructmol}, and LLaMo~\cite{park2024llamo} encode molecules with a 2D molecular graph encoder to capture the graph structural relations, while 3D-MoLM~\citep{li2024towards} uses 3D encoders to leverage the Euclidean distance information.
Despite the individual characteristics of each molecular modality, most works focus on dealing with specific molecular modalities.
CoLLaMo aims to leverage 1D, 2D, and 3D molecular information, enhancing generalizability across various tasks.

\section{Preliminaries}
\label{sec:3.1}

\noindent\textbf{Large Molecular Language Models.}
The goal of large molecular language models (LMLMs) is to enable large language models~(LLMs) to generate molecular instruction-following responses by incorporating molecular information into LLMs.
Formally, the input of LMLMs generally consists of two distinct types of tokens: molecule tokens $\boldsymbol{M}$ and instruction (text) tokens $\boldsymbol{T}$.
Given these input tokens, the language model predicts the response $\boldsymbol{Y}=\left\{y_i \right\}_{i=1}^K$ as:
\begin{equation}
\label{eq:lmlm}
p\left(\boldsymbol{Y}|\boldsymbol{M}, \boldsymbol{T}\right)=\prod_{i=1}^K p\left({y}_i | \boldsymbol{M}, \boldsymbol{T}, \boldsymbol{Y}_{<i}\right),
\end{equation}
where $\boldsymbol{Y}_{<i}$ denotes a generated token sequence until $i$-th token and $K$ is the number of generated tokens.

LMLMs generally consist of three modules: (i) molecular encoder, (ii) projector, and (iii) large language model~(LLM).
The molecular encoder learns representations of input molecules~(\textit{i.e.}, SELFIES, molecular graphs, molecular conformations).
Then, the projector transfers the molecular representations to molecular tokens comprehensible by the LLM.
Finally, the LLM generates the output response given the input molecular and instruction tokens.

\noindent\textbf{Molecular modalities.}
Molecules can be expressed with 1D molecular strings, 2D molecular graphs, and 3D molecular conformations, each offering distinct information.
Given a molecule, $m$, the 1D molecular string $\mathcal{S}_m$ is typically encoded by molecular descriptors such as SMILES~\citep{weininger1988smiles} and SELFIES~\citep{krenn2022selfies}, which provide a compact description of atoms and their connectivity in the string form.
The 2D molecular graph $\mathcal{G}_m = \left(\mathcal{V}_m, \mathcal{E}_m \right)$ represents the molecule’s topological structure, where atoms $\mathcal{V}_x$ are nodes and bonds $\mathcal{E}_x$ are edges.
The 3D molecular conformation $\mathcal{C}_m = \left(\mathcal{V}_m,\mathcal{P}_m\right)$ assigns spatial coordinates $\mathcal{P}_m$ to each atom, capturing stereochemical and dynamic properties critical for understanding molecules. 

\begin{table}
    \centering
    \renewcommand{\arraystretch}{0.93}
    {\setlength{\tabcolsep}{1.5pt}
    \resizebox{1.0\linewidth}{!}{
    \begin{tabular}{l|cccccccc}
    \toprule  
    \multirow{2}{*}{Molecular Modality} &\multirow{2}{*}{Captioning ($\uparrow$)}& Descriptive&  \multirow{2}{*}{LogP($\downarrow$)} & {TPSA} & {H-L Gap } & {SCF}\\
    &&QA($\uparrow$)&   &(\text{\AA}$^2$)($\downarrow$) & (eV)($\downarrow$)&($10^4$eV)($\downarrow$) \\
    \midrule
    1D String &25.40& 40.87&  1.31&38.43& 0.46 & 0.56 \\
    2D Graph &\textbf{32.14}& \textbf{42.74}& \textbf{0.87}&20.13& 0.31 &0.61  \\
    3D Conformation&30.72& 42.48& 0.93&\textbf{16.98}& \textbf{0.28} & \textbf{0.11}  \\
    \bottomrule
    \end{tabular}
    }
    \caption{Performance comparison of large molecular language models across three molecular modalities.}
    \label{tab:modality}
    }
\end{table}
\section{Impact of Molecular Modalities}
\label{sec:3.2}
Each molecular modality captures distinct facets of molecular structure that may influence the performance of molecule-related tasks.
To assess the impact of each modality on diverse tasks, we compare model performance using 1D, 2D, and 3D molecular representations.
We construct molecule tokens $\boldsymbol{M}$ differently according to the molecular modality and then feed them with instruction tokens $\boldsymbol{T}$ into the LLM to generate a response to the given tasks.

\noindent\textbf{LMLM architecture with 1D molecular modality.}
To use 1D molecule representation~(\textit{i.e.}, Mol-Instructions~\cite{fang2023mol}), the molecule tokens $\boldsymbol{M}_{\text{1d}}$ are defined as:
\begin{equation}
    \boldsymbol{M}_{\text{1d}} = \text{Token-Embed}\left(\mathcal{S}_m\right),
\end{equation}
where $\text{Token-Embed}\left(\cdot \right)$ denotes a token embedding function.
In our experiments, we adopt SELFIES as a molecular descriptor and employ a separate vocabulary based on the SELFIES alphabet instead of using the standard BPE tokenizer following existing works~\cite{pei2023biot5}.

\noindent\textbf{LMLM architecture with 2D/3D molecular modalities.}
While 1D molecular sequences are straightforward to embed using token embeddings, integrating 2D/3D molecular information into LLMs requires additional processing.
To address this, recent works~(\textit{e.g.}, MolCA~\cite{liu2023molca}, LLaMo~\cite{park2024llamo}, 3D-MoLM~\cite{li2024towards})  bridge the gap between 2D/3D molecules and natural language by adopting a learnable projector that converts 2D/3D molecular representations into token sequences:
\begin{equation*}
    \boldsymbol{M}_{\text{2d}} = \text{Proj}\left(g_{\text{2d}}\left(\mathcal{G}_m\right)\right), \ 
    \boldsymbol{M}_{\text{3d}} = \text{Proj}\left(g_{\text{3d}}\left(\mathcal{C}_m \right)\right)
\end{equation*}
where $g_{\text{2d}}, g_{\text{3d}}$ are 2D molecular graph and 3D molecular conformer encoder, respectively.
$\text{Proj}\left(\cdot \right)$ is a projector that aligns the 2D/3D encoder with the language models, enabling them to understand 2D/3D molecules.
We use pre-trained GIN \cite{xu2018powerful} for the 2D encoder and pre-trained UniMol \cite{zhou2023uni} for the 3D encoder, following previous works.
For the projector, we use a MLP similar to existing works~\cite{liu2024visual,cao2023instructmol}. 

\noindent\textbf{Evaluation tasks.}
We evaluate LMLMs across various tasks, including molecule captioning, descriptive property QA, and computed property QA.
For the computed property QA, we assess four key properties: LogP, TPSA, HOMO-LUMO gap, and SCF energy. 
We compare LMLMs that utilize different molecular modality-based tokens (1D, 2D, and 3D).
For each task, we compare models that utilize 1D, 2D, or 3D molecular modalities to analyze how modality-specific representations influence performance.

\noindent\textbf{Results.}
We report the experimental results in Table~\ref{tab:modality}. 
We observe several important trends.
3D modalities yield the best performance on coordinate-sensitive tasks, such as HOMO-LUMO gap and SCF energy prediction. This highlights the critical role of spatial information in tasks requiring accurate modeling of electronic structure.
2D graphs perform best on topology-driven tasks, such as LogP prediction, where subgraph patterns and bonding relationships are most informative.
1D SELFIES-based models generally underperform in comparison, suggesting that while sequence representations offer simplicity, they lack the fine-grained structural information required for more complex reasoning.
These results confirm that each modality contributes unique information, which is not fully captured by other modalities.
This underscores the importance of designing LMLMs that can flexibly integrate and reason over all three modalities to support a wide range of molecular tasks.
\section{CoLLaMo}
\label{sec:3.3}

\begin{figure*}[t!]
    \centering
    \includegraphics[width=0.95\textwidth]{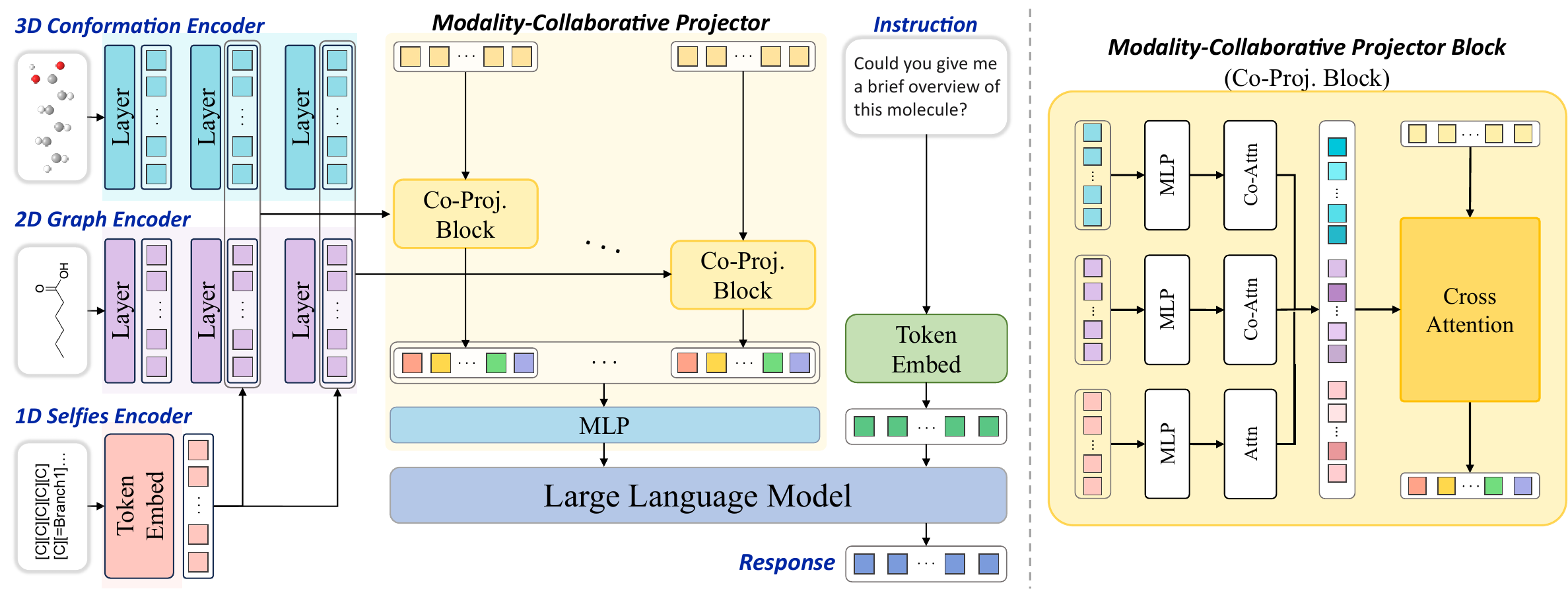}
    \caption{ 
    Overall architecture of CoLLaMo. Our framework, CoLLaMo, consists of encoders for 1D SELFIES, 2D graphs, and 3D conformations, a molecular modality-collaborative projector, and a large language model. Our framework first encodes 1D, 2D, and 3D molecular representations and then converts the encoded outputs into unified molecule tokens using the modality-collaborative projector. Finally, the large language model generates the response given the input molecule and instruction. 
    }
    \label{fig:main_figure}
\end{figure*}

Inspired by the observations in Section~\ref{sec:3.2}, we propose {CoLLaMo}, \textbf{Co}llaborative \textbf{L}arge \textbf{La}nguage model-based \textbf{Mo}lecular assistant that unifies the strength of different molecular modalities with a multi-level molecular modality-collaborative projector.
CoLLaMo consists of molecule encoders, the molecular modality-collaborative projector, and an LLM, which is depicted in Figure~\ref{fig:main_figure}.

\subsection{Molecular modality-collaborative projector}
The goal of {molecular modality-\textbf{co}llaborative \textbf{proj}ector~(Co-Proj.)} is to leverage 1D SELFIES, 2D graphs, and 3D conformations of the input molecule by encoding them under the shared molecule token space.
The Co-Proj. transforms the 1D, 2D, and 3D molecular representations, $\boldsymbol{Z}_{\text{1d}}=\text{Token-Embed}\left(\mathcal{S}_x\right) \in \mathbb{R}^{s \times d_1},\boldsymbol{Z}_{\text{2d}}=g_{\text{2d}}\left(\mathcal{G}_x \right) \in \mathbb{R}^{n \times d_2},\boldsymbol{Z}_{\text{3d}}=g_{\text{3d}}\left(\mathcal{C}_x \right) \in \mathbb{R}^{n \times d_3}$, into a fixed length of molecule token sequence $\boldsymbol{M}_{\text{unify}}$ to enable the LLM to understand the molecule.
To exploit hierarchical modality information, we adopt a multi-level fusion approach inspired by \citet{park2024llamo}.
Specifically, we first process the molecular representations with the attention module as $\hat{\boldsymbol{Z}}_{m}^{(l)} = \text{Attn}\left(\boldsymbol{Z}_{m}^{(l)}, \boldsymbol{Z}_{m}^{(l)}, \boldsymbol{Z}_{m}^{(l)} \right) + \boldsymbol{Z}_{m}^{(l)},$
where $\boldsymbol{Z}^{(l)}_m$ indicates hidden representations from $l$-th layer of $m$-D encoder and $\text{Attn}\left(Q,K,V\right)$ is the attention with query $Q$, key $K$, and value $V$.
For the 1D representation, we directly use the token embedding output for all layers.

\noindent\textbf{Relation-aware modality-collaborative attention.}
For 2D and 3D modalities, we adopt modality-collaborative attention instead of standard attention. 
The 2D graph and 3D conformation have the same atoms but different distinct relational structures~\cite{yu2024multimodal}.
To effectively integrate both 2D and 3D relational information, we design a collaborative attention~(Co-Attn) that captures their different relational information with the attention bias instead of the standard attention, which is formulated as:
\begin{equation}
\begin{split}
    &\text{Co-Attn}\left(Q, K, V\right) =\\
    &\text{softmax}\left(\frac{\left(Q W_Q \right)\left(K W_K \right)^\top}{\sqrt{d}} + \boldsymbol{\Phi}_{\text{2d}} + \boldsymbol{\Phi}_{\text{3d}} \right)V W_V,
\end{split}
\end{equation}
where $\boldsymbol{\Phi}_{\text{2d}}, \boldsymbol{\Phi}_{\text{3d}}$ are attention bias based on 2D and 3D structural information, respectively.

Specifically, we define $\boldsymbol{\Phi}_{\text{2d}}$ using the shortest path distance between atom pairs, a standard measure for capturing graph connectivity.
The shortest path distance-based attention bias $\boldsymbol{\Phi}_{\text{2d}}$ is defined as a learnable scalar value, where the value is determined by the shortest path between atoms for each pair of atoms.
For 3D structural information, we use Euclidean distance to leverage the spatial information in 3D space.
Following \citet{yu2024multimodal}, we first use the Gaussian Basis Kernel function~\citep{scholkopf1997comparing} to process the distance, ${\psi}_{(i,j)}^k = -\frac{1}{\sqrt{2\pi}\left\lvert \sigma^k\right\rvert}\exp\left(-\frac{1}{2}\left(\frac{\left\lVert \mathbf{r}_i-\mathbf{r}_j\right\rVert-\mu^k}{\left\lvert \sigma^k\right\rvert} \right)^2 \right)$, where $\mu^k, \sigma^k \in \mathbb{R}$ are learnable parameters.
Then, the 3D distance-based attention bias $\boldsymbol{\Phi}_{\text{3d}} \in \mathbb{R}^{n \times n}$ is calculated as $\boldsymbol{\Phi}_{\text{3d}} = \text{MLP}\left(\left[\Psi^1;\dots ; \Psi^K  \right] \right)$, where $\Psi^k$ indicates the processed distance matrix based on the $k$-th Gaussian kernel and $K$ is the number of Gaussian Basis kernel.

\noindent\textbf{Modality token unification via cross-attention.}
To produce a condensed and unified molecule tokens, we adopt cross-attention with the learnable tokens $\boldsymbol{P}^{(l)} = \left[\boldsymbol{p}^{(l)}_1; \dots; \boldsymbol{p}^{(l)}_b \right] \in \mathbb{R}^{d \times b}$ for $l=1, \dots, L$ to deal with an arbitrary number of atoms.
Specifically, the representations $\hat{\boldsymbol{Z}}_{\text{1d}}^{(l)}, \hat{\boldsymbol{Z}}_{\text{2d}}^{(l)}, \hat{\boldsymbol{Z}}_{\text{3d}}^{(l)}$ are unified into a fixed number of tokens $\hat{\boldsymbol{P}}^{(l)}$ as:

\begin{equation*}
\label{eq:temp}
    \hat{\boldsymbol{P}}^{(l)} =\text{Attn}\left(\boldsymbol{P}^{(l)}, \hat{\boldsymbol{Z}}_{\text{concat}}^{(l)}, \hat{\boldsymbol{Z}}_{\text{concat}}^{(l)} \right),
\end{equation*}
where $\hat{\boldsymbol{Z}}_{\text{concat}}^{(l)}$ is defined as $\left[\hat{\boldsymbol{Z}}_{\text{1d}}^{(l)};\hat{\boldsymbol{Z}}_{\text{2d}}^{(l)}; \hat{\boldsymbol{Z}}_{\text{3d}}^{(l)}\right]$.
We then construct unified molecule tokens $\boldsymbol{M}_{\text{unify}}\in \mathbb{R}^{d \times (b \cdot L)}$ by feeding tokens $\hat{\boldsymbol{P}}^{(l)}$ into MLP as: $\boldsymbol{M}_{\text{unify}} = \text{MLP}\left(\left[\hat{\boldsymbol{P}}^{(1)}; \dots ; \hat{\boldsymbol{P}}^{(L)} \right] \right)$.
This enables the molecule tokens to unify different molecular modalities in a shared semantic space, which promotes modality collaboration.

\noindent\textbf{Molecular modality dropout \& embedding.}
To prevent over-reliance on a specific modality, we apply modality dropout during pretraining by randomly masking each modality’s representation.
This encourages the model to learn more robust representations by leveraging the complementary information provided by the remaining modalities.
To further inform the model of the active modalities, we incorporate modality-specific learnable embeddings: $ \boldsymbol{P}^{(l)} = \boldsymbol{P}^{(l)}_{\text{base}} + \sum_{m \in \texttt{active}} \boldsymbol{P}^{(l)}_m,$
where $\texttt{active} \subseteq \{\text{1d}, \text{2d}, \text{3d} \}$.
For example, when the 2D molecular modality is dropped, we use the learnable query tokens  $\boldsymbol{P}^{(l)} = \boldsymbol{P}^{(l)}_\text{base}+ \boldsymbol{P}^{(l)}_\text{1d} + \boldsymbol{P}^{(l)}_\text{3d}$.
By jointly leveraging both modality dropout and modality embeddings, our approach enhances the model's generalization ability across tasks without being overly dependent on specific molecular modality.
\subsection{Training CoLLaMo}
\label{sec:3.4}

\begin{table*}[!t]
    
    \centering
    \renewcommand{\arraystretch}{1.05}
    \resizebox{0.8\textwidth}{!}
    {
    \begin{tabular}{c c|c c|c c|c c}
        \toprule
        \multirow{2}{*}{\textbf{Model}} & \multirow{2}{*}{\textbf{LLM}}  & \multicolumn{2}{c|}{\textbf{Molecule Captioning}} & \multicolumn{2}{c|}{\textbf{IUPAC Prediction}} & \multicolumn{2}{c}{\textbf{Motif Counting}} \\
         & &BLEU~($\uparrow$) & METEOR~($\uparrow$) & BLEU~($\uparrow$) & METEOR~($\uparrow$) & MAE~($\downarrow$) & Validity~($\uparrow$) \\
        \midrule
        \midrule
         \multicolumn{1}{l}{GPT-4} & GPT-4  & 0.8 & 16.7 & 28.3 & 50.4 &5.78& 100\%\\
         \multicolumn{1}{l}{GPT-4 (ICL)} & GPT-4  & 27.0 & 52.2 & 51.7 & 62.7 & 2.51& 100\%\\
         \multicolumn{1}{l}{GPT-4o} & GPT-4o  & 1.6 & 15.6 & 11.2 &37.2&4.21& 95\%\\
         \multicolumn{1}{l}{GPT-4o (ICL)} & GPT-4o  & 31.1 & 57.1 & 39.4 & 55.7&2.45& 100\%\\
         \multicolumn{1}{l}{o1-mini} & o1-mini  & 1.8 & 16.6 & 2.4 & 26.5 &9.62& 100\%\\
         \multicolumn{1}{l}{o1-mini (ICL)} & o1-mini & 27.2 & 53.0 & 29.4 &50.5&2.72& 100\%\\
         \midrule
         \multicolumn{1}{l}{LLaMA2} & LLaMA2-7B  & 0.0 & 14.1 & 0.0& 0.4&N/A$^*$ & --\\
         \multicolumn{1}{l}{Mol-Instructions} & LLaMA2-7B  & 14.3 & 25.4 & -- & -- &-- & --\\
         \multicolumn{1}{l}{LLaMo} & LLaMA2-7B & 38.9 & 67.1 & 56.0 & 73.2 & -- & --\\
        \rowcolor{Aquamarine!20}
         \multicolumn{1}{l}{\textbf{CoLLaMo (Ours)}} & LLaMA2-7B & \textbf{44.5} & \textbf{70.5} & \textbf{59.8} & \textbf{77.0}&\textbf{1.31}& \textbf{100\%}\\
        \hline
    \end{tabular}
}
    \caption{ {Evaluation results of generalist models on molecule captioning, hallucination, IUPAC prediction, and motif counting. 
    ICL denotes in-context learning.
    $*$ The result is not available since LLaMA2 fails generating numerical outputs.}
    }
    \label{tab:generalist}
\end{table*}

\begin{table*}
\centering
\scriptsize
\setlength{\tabcolsep}{1.3pt}

\begin{tabular}{l| >{\centering\arraybackslash}m{1.3cm} >{\centering\arraybackslash}m{1.2cm} | >{\centering\arraybackslash}m{1.2cm} >{\centering\arraybackslash}m{0.8cm} >{\centering\arraybackslash}m{0.8cm} >{\centering\arraybackslash}m{0.8cm}  >{\centering\arraybackslash}m{0.8cm} >{\centering\arraybackslash}m{0.8cm} >{\centering\arraybackslash}m{0.8cm} >{\centering\arraybackslash}m{0.8cm} >{\centering\arraybackslash}m{0.8cm} >{\centering\arraybackslash}m{0.8cm} >{\centering\arraybackslash}m{0.8cm} >{\centering\arraybackslash}m{0.8cm}}

\toprule
 \multirow{3}{*}{Model} & \multicolumn{2}{c|}{\textbf{Descriptive Property QA}} & \multicolumn{12}{c}{\textbf{Computed Property QA}}\\
  & \multirow{2}{*}{BLEU} & \multirow{2}{*}{METEOR}&  \multicolumn{2}{c}{\textcolor{black}{LogP}} &\multicolumn{2}{c}{\textcolor{black}{TPSA (\text{\AA}$^2$)}} &\multicolumn{2}{c}{\textcolor{black}{HOMO (eV)}} & \multicolumn{2}{c}{\textcolor{black}{LUMO (eV)}} & \multicolumn{2}{c}{\textcolor{black}{H-L Gap (eV)}} & \multicolumn{2}{c}{\textcolor{black}{SCF ($10^4$eV)}} \\
  &  &  &  MAE & Validity &  MAE & Validity & MAE & Validity & MAE & Validity & MAE & Validity & MAE & Validity \\
\midrule
Uni-Mol~(Non-LM) & -- & -- & 0.59& \textbf{100\%} & 13.48 & \textbf{100\%} & 0.32 & \textbf{100\%} & 0.35 & \textbf{100\%} & 0.21 & \textbf{100\%} & 0.45 & \textbf{100\%} \\
\hline
\rowcolor{lightgray}
\textbf{\textit{Specialist}}&&&&&&&&&&&&&& \\
{Llama2-7B}        &     {23.24} &  {46.87} & 1.45 & 95\% & 15.87 & 92\% & 1.24 & 96\% & 1.04 & 95\% & 0.88 & 92\% & 0.70 & 99\% \\
 {2D-MoLM}        &    {{25.09}}  &  {50.92} & 0.88 & 96\% & 13.52 & 92\% & 0.92 & 98\% & 0.80 & 96\% & 0.67 & 93\% & 0.71 & 99\% \\
{3D-MoLM$\dag$}        &     {24.47} &  {{51.33}} & 0.95& 96\% & 10.26 & 94\% &  {0.45} & 98\% & {0.36} & 96\% & {0.41} & 94\% & {0.39} & 99\% \\
{3D-MoLM}        &     {26.13} &  {52.15} & 0.66& 97\% & {9.71} & 93\% & 0.26 & 97\% & 0.25 & 94\% & 0.28 & 94\% & 0.35 & 99\% \\
\rowcolor{Aquamarine!20}
\textbf{CoLLaMo~({Ours})} & \textbf{45.62} & \textbf{60.06} & \textbf{0.37} & \textbf{100\%} & \textbf{4.93} & \textbf{100\%} &\textbf{0.08} & \textbf{100\%} & \textbf{0.08} & \textbf{100\%} & \textbf{0.09} & \textbf{100\%} & \textbf{0.02} & \textbf{100\%} \\
\hline
\hline
\rowcolor{lightgray}
\textbf{\textit{Generalist}}&&&&&&&&&&&&&& \\
Llama2-7B*        &   {21.16}  & {43.17} & 2.10 & 85\% & 27.11& 76\%& 2.87 & 70\% & 1.89 & 71\% & 1.86 & 70\% & 3.84 & 23\% \\
{Llama2-7B}        &     {22.81} &  {46.39} & 1.78 & 93\% & 17.07 & 90\% & 1.89 & 90\% & 1.26 & 90\% & 1.25 & 91\% & 0.87 & 99\% \\
 {2D-MoLM}        &    {{24.57}}  &  {50.08} & 1.36 & 94\% & 12.47 & 89\% & 1.52 & 93\% & 1.13 & 92\% & 1.09 & 88\% & 0.96 & 99\% \\
{3D-MoLM$\dag$}        &     {24.44} &  {{50.81}} & 0.92& 92\% & 11.14 & 92\% &  {0.65} & 94\% & {0.41} & 92\% & {0.55} & 89\% & {0.49} & 99\% \\
{3D-MoLM}        &     {26.08} &  {51.93} & 0.78& 95\% & 10.90 & 90\% & 0.35 & 95\% & 0.36 & 93\% & 0.32 & 90\% & 0.38 & 98\% \\
\rowcolor{Aquamarine!20}
\textbf{CoLLaMo~({Ours})} & \textbf{45.62} & \textbf{59.89} & \textbf{0.63} & \textbf{100\%} & \textbf{7.79} & \textbf{100\%} &\textbf{0.13} & \textbf{100\%} & \textbf{0.14} & \textbf{100\%} & \textbf{0.13} & \textbf{100\%} & \textbf{0.09} & \textbf{100\%} \\
\hline
\end{tabular}
\caption{Experimental results for descriptive property QA and computed property QA. For computed property QA, MAE is used for the evaluation metric. The validity is also reported since LMs sometimes fail to generate valid numerical responses.
* represents the model without fine-tuning. $\dag$ denotes the variant of 3D-MoLM that is pretrained using original PubChem text without GPT-3.5 enrichment.
}
\label{tab:qa}
\end{table*}
 
Following most existing works, our CoLLaMo is trained in multiple stages.

\noindent\textbf{Stage 1. Pretraining for molecule-language alignment.}
The goal of the first stage is to align molecule encoders with a large language model.
We train our Co-Proj. and molecule encoders while the LLM is being frozen.
In this stage, we use two types of molecule-caption pair datasets: a molecule-caption pair dataset from \citet{fang2023mol} and the GPT-enriched data presented in \citet{li2024towards}.

\noindent\textbf{Stage 2. Molecule-language instruction-tuning.}
The second stage aims to improve the molecule-language instruction-following capabilities of LLMs and enable them to understand molecules.
We train both the Co-Proj. and LLM equipped with LoRA~\citep{hu2021lora} while keeping the molecule encoders frozen.
In this stage, we employ a diverse set of datasets with various instructions: training splits of molecule captioning, descriptive property QA, computed property QA, motif counting, and IUPAC prediction.

\section{Advancing Molecule-Aware Evaluation}
\paragraph{Hallucination assessment.}
Although recent large molecule-language models (LMLMs)~\cite{park2024llamo,liu2023molca,li2024towards} have shown progress in generating molecule-related descriptions, their outputs often contain factual inaccuracies or spurious content.
Despite the importance of the hallucination of LMLMs, previous studies have underexplored the evaluation metrics for the hallucination.
So, we introduce two quantitative metrics specifically designed to evaluate the factual consistency of LMLM outputs.

Inspired by CHAIR metric \cite{rohrbach2018object}, we propose \textbf{CHARM} (\textbf{C}aption \textbf{H}allucination \textbf{A}ssessment with \textbf{R}elevance to \textbf{M}olecule), a metric designed to assess hallucination in molecule-language models.
CHARM quantifies the proportion of mentioned molecular entities that are factually incorrect or not grounded in the input molecule.
Specifically, we first extract molecular entities by applying BERN2~\cite{sung2022bern2}, a widely-used and effective Named Entity Recognition (NER) system.
To ensure the reliability of the extracted entities, we filtered out low-confidence predictions and only retained entities with a confidence score of 0.9 or higher.
With extracted entities, we calculate metrics as follows:
\begin{equation}
    \mathrm{CHARM}=\frac{|\text{\{hallucinated entities\}}|}{|\{\text{all mentioned entities}\}|},
\end{equation}
This value represents the proportion of hallucinated entities among all entities mentioned.
As CHARM corresponds to a precision-oriented view of factuality, we further propose \textbf{RCHARM}, a complementary, recall-oriented metric that captures the proportion of relevant entities from the ground truth that are omitted in the generated output:
\begin{equation}
    \mathrm{RCHARM}=\frac{|\text{\{not mentioned g.t. entities\}}|}{|\{\text{all g.t. entities}\}|}.
\end{equation}
Together, CHARM and RCHARM enable a more comprehensive and balanced assessment of hallucination in molecule-language models.

\paragraph{LLM-based molecule description assessment.}
Traditional evaluation metrics, such as BLEU and ROUGE, are occasionally inadequate for assessing molecular descriptions, where ground-truth references may be underspecified or lack molecular precision.
To overcome these limitations, we adopt a large language model (LLM)-based evaluation approach, following recent trends in LLM-as-a-judge setups~\citep{zheng2023judging}.
Specifically, we employ GPT-4o to assess the quality of generated molecular descriptions from CoLLaMo and baseline models.
The evaluation prompt is structured to include: (i) the input molecule in SELFIES format, (ii) the ground-truth description, and (iii) the model-generated output.
The LLM judge is asked to score the response based on two criteria: factual informativeness and alignment with the ground truth.
Based on criteria, the model assigns a human-likeness score on a 0–5 scale.

\section{Experiments}
\subsection{Experimental Settings}
\noindent\textbf{Benchmarks.}
To validate the effectiveness of our CoLLaMo, we compare the performance of baseline models on diverse tasks such as  \textbf{1)} molecule captioning, \textbf{2)} molecule hallucination assessment, \textbf{3)} descriptive property QA, \textbf{4)} computed property QA, \textbf{5)} IUPAC prediction, and \textbf{6)} motif counting.

\noindent\textbf{Implementation details.}
We utilize the LLaMA 2 Chat 7B model as our base language model.
For the analysis, including the ablation studies, we employ a \textbf{short training schedule} consisting of 64k iterations of pre-training followed by 150k iterations of instruction tuning. 
For the final models, we adopt an \textbf{extended training schedule} with 64k iterations of pre-training and 900k iterations of instruction tuning.
We conduct experiments on two variants of CoLLaMo: generalist, which is a single model trained on all the tasks, and specialist, which is trained for each task following other works~\cite{li2024towards}.

\subsection{Experimental Results}

\begin{table}[t!]
    \centering
    \resizebox{\columnwidth}{!}{
    \renewcommand{\arraystretch}{1.0} 
    \setlength{\tabcolsep}{4pt}
    \begin{tabular}{l|cc|c}
        \toprule
        \multirow{2}{*}{\textbf{Model}} & \multicolumn{2}{c|}{\textbf{Molecular Entity Hallucination}} & \textbf{LLM Score} \\
        & $\text{CHARM}$~($\downarrow$) & $\text{RCHARM}$~($\downarrow$) & Avg. Score ~($\uparrow$) \\
        \midrule
        GPT-4 &  98.4 & 94.3 & 1.95   \\
        GPT-4 (ICL) &  77.1 & 76.9  & 2.04  \\ 
        GPT-4o  & 99.0 & 98.4  & 1.99  \\
        GPT-4o (ICL) & 74.0 & 75.1  & 2.15  \\
        o1-mini &  99.1 & 98.8 & 2.17   \\
        o1-mini (ICL) & 83.5 & 82.7 & 2.01  \\ 
        LLaMo & 64.7 & 67.2 & 2.17  \\
        \rowcolor{Aquamarine!20}
        \textbf{CoLLaMo~(Ours)} & \textbf{58.5} & \textbf{59.9} & \textbf{2.52}  \\
        \hline
    \end{tabular}}
    \caption{
    Evaluation with our evaluation protocols, such as CHARM/RCHARM and GPT-based LLM score. GPT-4o is used as a judge.
    }
    \label{tab:evaluation}
\end{table}

\begin{table}
    \centering
    \resizebox{\columnwidth}{!}{
    \renewcommand{\arraystretch}{1.0} 
    \setlength{\tabcolsep}{4pt} 
    \begin{tabular}{cccc|cccc}
        \toprule
        \multicolumn{3}{c}{Mol. modality} & Proj. 
 & Captioning & Motif Count & \multicolumn{1}{c}{Desc.  QA} & H-L Gap \\
 {\textbf{1D}} & {\textbf{2D}} & {\textbf{3D}} & \textbf{C.P}
   &BLEU~($\uparrow$) &MAE~($\downarrow$) & BLEU~($\uparrow$)& MAE~($\downarrow$) \\
        \midrule
    \checkmark & {} & {} & {} & 33.2 & 1.82 & 42.9 & 0.31  \\
    {} & \checkmark & {} & {} &  38.0 & \textbf{1.41} & 43.9 & 0.27  \\
    {} & {} & \checkmark & {} & 35.8 & 1.65 & 42.6 & 0.23  \\
    \checkmark & \checkmark & \checkmark & {} & 35.7 & 1.59 & 43.6 & 0.21  \\
    \rowcolor{Aquamarine!20}
    \checkmark & \checkmark & \checkmark & \checkmark & \textbf{40.1} & 1.43 & \textbf{44.8} & \textbf{0.18}  \\
        \hline
    \end{tabular}}
    \caption{
    {Ablation study on molecular modality. C.P denotes molecular modality-\textbf{c}ollaborative \textbf{p}rojector.}
    }
    \label{modality ablation}
\end{table}

\begin{table}[t!]
    \centering
    \resizebox{\columnwidth}{!}{
    \renewcommand{\arraystretch}{1.0} 
    \setlength{\tabcolsep}{4pt}
    \begin{tabular}{l|cccc}
        \toprule
        \multirow{2}{*}{\textbf{Component}} & Captioning & Motif Count & \multicolumn{1}{c}{Desc.  QA} & H-L Gap \\
      &BLEU~($\uparrow$) &MAE~($\downarrow$) & BLEU~($\uparrow$)& MAE~($\downarrow$) \\
        \midrule
        w/o Co-Attention &  36.6 & 1.44 & 43.9 & 0.19   \\
        w/o Modality embedding &  38.9 & 1.51 & 44.3 & 0.19  \\ 
        w/o Modality dropout &  37.8 & 1.49 & 44.5 & \textbf{0.18}  \\
        \rowcolor{Aquamarine!20}
        \textbf{CoLLaMo~(Ours)} &  \textbf{40.1} & \textbf{1.43} & \textbf{44.8} & \textbf{0.18}  \\
        \hline
    \end{tabular}}
    \caption{
    Ablation study on components of CoLLaMo. 
    }
    \label{tab:component}
\end{table}

\begin{table}[t]
    \centering
    \resizebox{\columnwidth}{!}{
    \renewcommand{\arraystretch}{1.0} 
    \setlength{\tabcolsep}{4pt} 
    \begin{tabular}{lc|cccc}
        \toprule
  \multirow{2}{*}{Co-Proj.} & Adopted  &\multicolumn{1}{c}{Captioning}& Motif Count & Desc. QA & H-L Gap\\
   &Mol. modality & BLEU & MAE & BLEU & MAE\\
        \midrule
    & {1D}  & 2.4 & N/A & 12.8 & N/A \\
   w/o Co-Proj. & {1D+2D} & 3.3  & N/A &13.8 & N/A \\
    & {1D+2D+3D}  & 35.7 & 1.59 & 43.6 & 0.21 \\
    \hline
     \cellcolor{Aquamarine!20}& {1D} \cellcolor{Aquamarine!20} & \cellcolor{Aquamarine!20}28.0 & \cellcolor{Aquamarine!20}1.94 & \cellcolor{Aquamarine!20} 42.4 & \cellcolor{Aquamarine!20} 0.31\\
   \textbf{w/ Co-Proj.~(Ours)} \cellcolor{Aquamarine!20}& {1D+2D} \cellcolor{Aquamarine!20}& \cellcolor{Aquamarine!20}39.1 & \cellcolor{Aquamarine!20}\textbf{1.43} & \cellcolor{Aquamarine!20}44.0 & \cellcolor{Aquamarine!20}0.24 \\
     \cellcolor{Aquamarine!20}& {1D+2D+3D} \cellcolor{Aquamarine!20}& \cellcolor{Aquamarine!20}\textbf{40.1}  & \cellcolor{Aquamarine!20}\textbf{1.43} & \cellcolor{Aquamarine!20}\textbf{44.8}& \cellcolor{Aquamarine!20}\textbf{0.18} \\
        \hline
    \end{tabular}}
    \caption{
    Experimental results under \textbf{missing modality settings}. Adopted Mol. modality denotes molecular modality employed during the inference phase. Co-Proj. denotes molecular modality-collaborative projector.
    }
    \label{tab:missing_modalities}
\end{table}


\noindent\textbf{Molecule captioning.}
We evaluate CoLLaMo and baselines on molecule captioning in Table~\ref{tab:generalist}.
The table shows the superiority of CoLLaMo on molecule captioning.
In comparison to GPT-based models such as \texttt{GPT-4}, \texttt{GPT-4o}, and \texttt{o1-mini}, our CoLLaMo achieves the best performance.

\noindent\textbf{IUPAC prediction and motif counting.}
We provide additional experimental results on IUPAC name prediction and motif counting tasks in Table~\ref{tab:generalist}.
Our CoLLaMo consistently outperforms GPT-based models across both tasks with its effective molecular understanding capabilities.

\begin{figure*}[t!]
    \centering
    \includegraphics[width=0.95\textwidth]{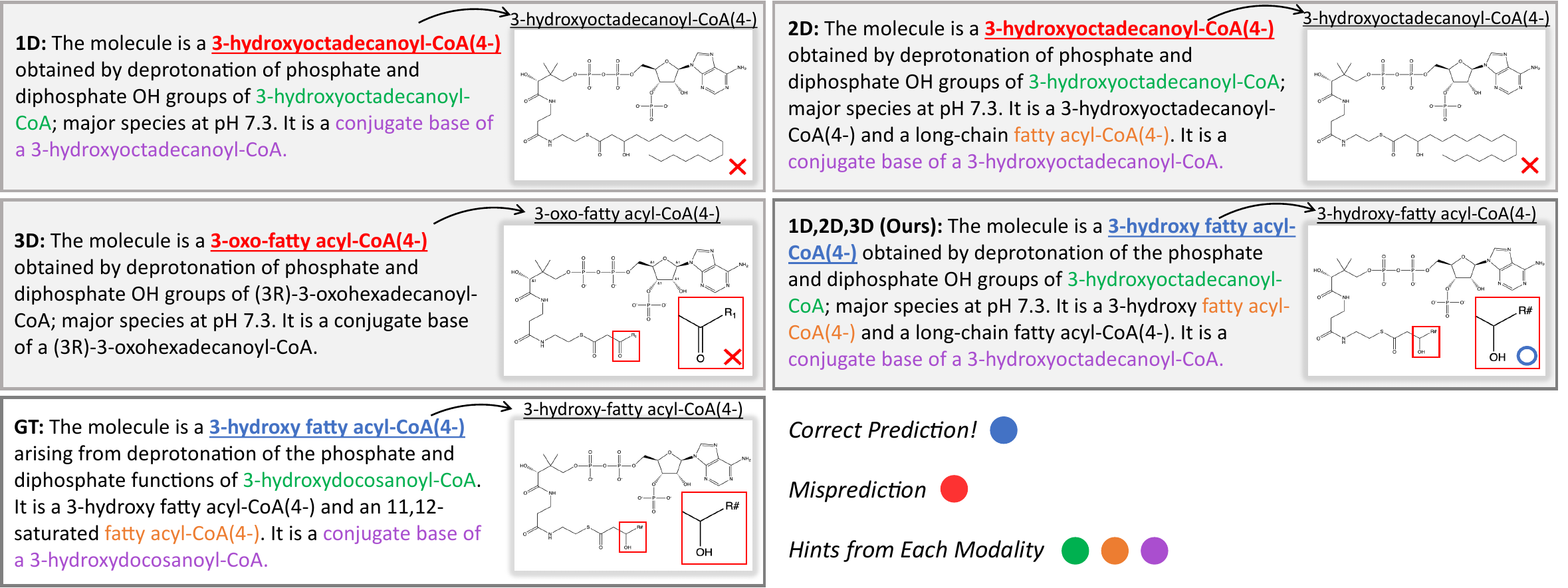}
    \caption{Examples of molecule captioning given the molecule (CHEBI:65102).
    The molecule structures of underlined texts in generated descriptions are illustrated in the white box.} 
    \label{fig:qual_figure}
\end{figure*}

\noindent\textbf{Molecular property QA.}
Table~\ref{tab:qa} presents the experimental results on open-text descriptive and computed property QA tasks.
From the table, our CoLLaMo outperforms the baselines using 2D molecular graphs~(2D-MoLM) or 3D conformations~(3D-MoLM) in both \textit{Generalist} and \textit{Specialist} settings.
In particular, our CoLLaMo achieves a significant performance improvement of \textcolor{black}{19.54 in BLEU} and \textcolor{black}{7.96 in METEOR} compared to 3D-MoLM on the descriptive property QA task in the \textit{Generalist} setting.
These results highlight that 1D, 2D, and 3D modalities have effectively collaborated.
Additionally, CoLLaMo surpasses molecule-specialized non-LM model Uni-Mol on all properties in the \textit{Specialist} setting, underscoring its ability to combine the contextual knowledge from diverse molecular modalities.

\noindent\textbf{Molecular Entity hallucination.}
We evaluate the extent of hallucination using our proposed metrics, CHARM and RCHARM in Table~\ref{tab:evaluation}.
The table shows that CoLLaMo outperforms GPT-based models and LLaMo on both CHARM and RCHARM metrics, which demonstrates that CoLLaMo successfully collaborates with diverse molecular modalities and complements information with each other.

\noindent\textbf{LLM-based evaluation.}
To further validate the quality of generated outputs, we employed an LLM-as-a-judge setup using GPT-4o to evaluate responses from CoLLaMo and baseline models.
The average human-likeness scores (scale: 0–5) are summarized in Table~\ref{tab:evaluation}.
From the table, CoLLaMo achieves the highest score, demonstrating its superior molecule understanding and generation quality from both automatic and LLM-based evaluations.

\subsection{Analysis}
\noindent\textbf{Impact of molecular modalities and the molecular modality-collaborative projector.}
To assess the contribution of different molecular modalities, we evaluate the model using 1D only, 2D only, 3D only, and unification of 1D, 2D, and 3D molecular modalities with and without the molecular modality-collaborative projector~(C.P.) in Table~\ref{modality ablation}.
For the model without C.P., the cross-attention is separately applied for each molecular modality.
The results demonstrate that integrating 1D, 2D, and 3D modalities through the modality-collaborative projector consistently improves the performance across all tasks except motif counting.

\noindent\textbf{Effectiveness of components in CoLLaMo.}
Table~\ref{tab:component} shows the results of ablation studies on the key components of CoLLaMo, including Co-Attention, modality dropout, and modality embedding.
The experimental results reveal that all components contribute to the performance improvement of our CoLLaMo.
Notably, when comparing CoLLaMo with and without Co-Attention, we observe a significant performance improvement of 3.5 in the BLEU score on the molecule captioning task.
Additionally, the inclusion of Co-Attention yields substantial gains in performance, highlighting the importance of this regularization technique.

\noindent\textbf{Robustness under missing molecular modality.}
We further evaluate CoLLaMo's robustness in scenarios with incomplete molecular modalities by comparing the model with and without a molecular collaborative-modality projector~(Co-Proj.), as presented in Table~\ref{tab:missing_modalities}.
The robustness is assessed by utilizing an incomplete set of molecular modalities during the inference phase.
We note that all molecular modalities are used in the training phase.
The results reveal that even when only a single modality is available during inference, CoLLaMo maintains strong performance, demonstrating its robustness, while the model without Co-Projector is significantly degraded.
Since CoLLaMo embeds different molecular modalities into shared representations, it preserves molecule information given the incomplete molecular modalities.
These findings highlight CoLLaMo’s capacity to generalize well in real-world scenarios, where modality information may be incomplete or unavailable.

\noindent\textbf{Qualitative analysis.}
Figure~\ref{fig:qual_figure} presents a comparison between the ground truth (GT) caption and the captions generated by models trained with individual molecular modalities (1D string, 2D graph, 3D conformation) and all modalities combined.
As shown in the figure, CoLLaMo generates the most accurate molecular description, leveraging all molecular modalities.
The GT caption describes the molecule with ``\textcolor{black}{3-hydroxy fatty acyl-CoA(4-)}”.
While the model with a single molecular modality wrongly explains the molecule, CoLLaMo with all modalities generates an accurate description with ``3-hydroxy fatty acyl-CoA(4-)” by successfully integrating the information from each modality.

\section{Conclusion}
We propose CoLLaMo, a molecular modality \textbf{Co}llaborative projector-based \textbf{L}arge \textbf{La}nguage model-based \textbf{Mo}lecular assistant, which integrates 1D, 2D, and 3D modalities under a unified molecule token space.
We also present new measurements to assess the hallucination in the outputs generated by the language models, addressing the limitations of existing measurements.
Our experiments demonstrate the effectiveness of CoLLaMo, outperforming GPT-based models such as GPT-4, GPT-4o, and o1-mini on diverse tasks, including molecule captioning, IUPAC prediction, and motif counting.
In particular, our CoLLaMo shows good experimental results under missing modality settings, highlighting that it effectively integrates 1D, 2D, and 3D modalities.

\section*{Acknowledgements}
This research was supported by the ASTRA Project through the National Research Foundation (NRF) funded by the Ministry of Science and ICT (No. RS-2024-00439619).

\bibliography{aaai2026}

\end{document}